\documentclass[11pt]{article}

\usepackage[final]{acl}

\usepackage{times}
\usepackage{latexsym}
\usepackage[T1]{fontenc}
\usepackage[utf8]{inputenc}
\usepackage{microtype}
\IfFileExists{inconsolata.sty}{\usepackage{inconsolata}}{}
\usepackage{graphicx}
\usepackage{booktabs}
\usepackage{amsmath}
\usepackage{amssymb}
\usepackage{multirow}
\usepackage{xcolor}
\usepackage{enumitem}
\setlist{nosep, itemsep=2pt, topsep=3pt, parsep=0pt}

\title{Do All LLMs Know When They're Being Harmful? \\
A Reproducibility Study of Latent-Space Safety Probes Across Model Families}

\author{
  Alizishaan Khatri \\
  Wrynx Inc \\
  \texttt{research@wrynx.com} \\
  \And
  Dun Li Chan \\
  INTI International College Penang \\
}

\begin{document}
\maketitle

\begin{abstract}
\citet{khatri2026safety} show that lightweight MLP probes on final-layer
activations of a single 8B model (\textsc{LLaMA-3.1-8B}) detect harmful
prompts at F1 competitive with guard models 1000$\times$ larger, using
one probe per benchmark. We reproduce this pipeline end-to-end and extend it
along two axes the original study leaves open. First, we test whether the
result generalizes across other model \emph{architecture and scale} by training
identical probes on activations from models like Gemma-4-E4B, Mistral-7B-v0.3, and Qwen2-7B, using the three benchmarks
(WildJailbreak, BeaverTails, AEGIS~2.0). Second, we test how much of the
reported performance is affected by \emph{non-determinism} during inference by repeating extraction under five random seeds and measuring the variance of F1 scores.  Our results reproduce the original LLaMA model benchmarks within 0.37 percentage points of the original F1 scores (and within 0.2 points on BeaverTails). We find that the original MLP probe architecture extends to other model families with F1 scores within a point of the values reported for LLaMA-3.1-8B. Our experiments varying seed values reveals an interesting observation - final token latent vectors remained the same for all tested architectures irrespective of the seed values used.
\end{abstract}

\section{Introduction}
\label{sec:intro}

Runtime safety moderation for large language models (LLMs) is dominated by
\emph{external guardrails}: separate classifier models that screen prompts
and responses \citep{inan2023llama, zeng2024shieldgemma, han2024wildguard}.
\citet{khatri2026safety} argue that this architecture is unnecessarily
expensive, since safety-relevant information is already linearly decodable
from an LLM's own hidden states \citep{zou2023representation,
saglam2025large, chen2025towards}. They support this claim with a single
model family (\textsc{LLaMA-3.1-8B}) and a single extraction configuration
(final layer, last-token after prefill), reporting
F1 scores of 99.1\%, 82.7\%, and 83.5\% on WildJailbreak, BeaverTails, and
AEGIS~2.0 respectively \citep{jiang2024wildteaming, ji2023beavertails,
ghosh2025aegis}.

This leaves two questions the original paper does not address, and which
are central to whether latent-space probing is a viable general-purpose
alternative to external guardrails:

\begin{enumerate}
  \item \textbf{Generality across architectures.} Does the finding that
  ``safety lives in a linearly separable subspace'' hold for models trained
  with different architectures, pretraining data, alignment recipes, and hidden
  dimensionality, or is it an artifact of \textsc{LLaMA-3.1-8B}'s specific
  alignment procedure?
  \item \textbf{Sensitivity to non-determinism.} The reported metrics come
  from a single extraction-and-training run per dataset. Activation
  extraction involves stochastic components (sampling
  order, weight initialization,
  non-deterministic GPU kernels, etc). In a practical setting, one probe would be deployed one model per category. However, a user must be able to use different seed values for the same model. How sensitive are the probe and guard-model baselines in Table~3 of \citet{khatri2026safety} to non-determinism of latent states during inference?
\end{enumerate}

We address both questions directly. We reimplement the two-stage pipeline
of \citet{khatri2026safety} (activation extraction, then MLP probe
training) and (1) apply it unchanged to additional open-weight model
families at comparable parameter counts, and (2) repeat
extraction under multiple seeds for each model/ data set pair to
report probe variance along with point estimates. This positions our contribution
squarely within The 9th BlackboxNLP Workshop Special Track: Reproducibility and Reliability in Interpretability Analyses: we apply the original method
without modification to previously untested models and add the statistical
controls (multi-seed variance) the original study omits.

\paragraph{Contributions.}
\begin{itemize}
  \item A faithful reproduction of the \citet{khatri2026safety} probe
  pipeline, validated against the original paper's reported numbers on
  \textsc{LLaMA-3.1-8B}.
  \item Empirical benchmarks for three additional model families,
  testing whether probe performance and probe-vs-guard-model competitiveness
  hold outside the original model family.
  \item A variance analysis across five independent extraction runs per model/dataset pair, reporting standard deviation of F1 score, and quantifying
  what fraction of the original paper's reported margin over baseline guard
  models survives this noise.
  \item Released code, activations, and per-run results.
\end{itemize}

\section{Related Work}
\label{sec:related}

\paragraph{External guardrails.} Production LLM systems commonly rely on
dedicated guard models such as Llama Guard \citep{inan2023llama},
ShieldGemma \citep{zeng2024shieldgemma}, Latent Guard \citep{zhao2025latentguard}, and WildGuard
\citep{han2024wildguard} to screen prompts and responses \citep{dong2024building}. These add inference-time latency and remain
vulnerable to attacks that exploit the gap between what the guard sees and
what happens inside the primary model \citep{hung2025attention}.

\paragraph{Latent safety representations.} A growing line of work argues
that safety-relevant concepts are linearly represented in hidden LLM states 
\citep{zou2023representation, chen2025towards, chia2025probing}, and that
lightweight probes on these representations can match or approach
external classifiers at a fraction of the parameter count
\citep{saglam2025large, cunningham2026constitutional, zhang2024llmscan}.
Closely related, \citet{zhao2025latentguard} show that harmfulness and
refusal occupy separate directions in latent space, and that only the
harmfulness direction, not the refusal direction, governs the model's
judgment of harm. This distinction is directly relevant here, since the
probes we study target harmfulness rather than refusal.
\citet{khatri2026safety}, the paper we reproduce and extend, is the most
direct precedent: a 13.9M-parameter MLP probe on \textsc{LLaMA-3.1-8B}
final-layer activations, evaluated on WildJailbreak, BeaverTails,
and AEGIS~2.0. Concurrent work extends the cross-architecture question
we ask in Section~\ref{sec:ext-arch}: \citet{geometric2026harmful} show
that harmful intent is linearly separable from residual-stream activations
across 12 models spanning four architectural families and three alignment
variants (base, instruction-tuned, and abliterated), including cases where
the refusal mechanism has been surgically removed, evidence that harm
recognition and refusal generation are separable mechanisms at the
representation level, consistent with \citet{zhao2025latentguard} above.
Our contribution focuses on cross-architecture generalization and
run-to-run variance rather than head-to-head latency comparisons.

\paragraph{Reproducibility and robustness in interpretability.} Our
extensions follow the spirit of recent calls for stress-testing
interpretability claims across models and against random/non-deterministic
baselines rather than reporting single-run point estimates on a single
model \citep{albrethsen2026deepcontext}. We are not aware of prior work
that specifically tests cross-architecture generalization or extraction
non-determinism for latent safety probes.

\section{Methodology}
\label{sec:method}

\begin{figure}[t]
  \centering
  \includegraphics[width=0.90\columnwidth]{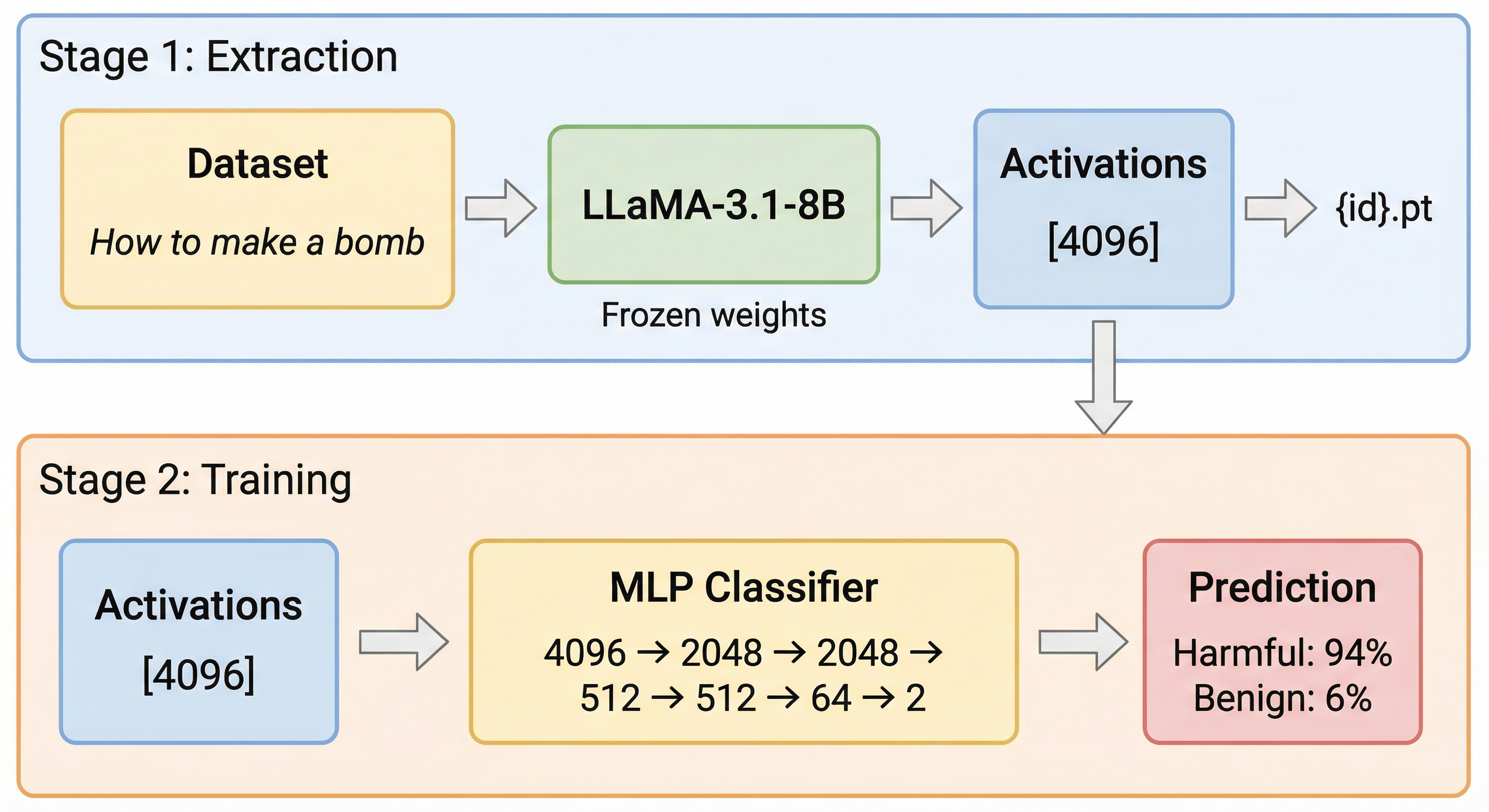}
  \caption{Reproduced extraction-and-probing pipeline. Only the backbone
  $\mathcal{M}$ and the random seed vary across our experiments.}
  \label{fig:pipeline}
\end{figure}

We reuse the two-stage pipeline of \citet{khatri2026safety} unchanged
(Figure~\ref{fig:pipeline}) and vary only (a) the backbone LLM from which
activations are extracted, and (b) the random seed governing extraction
order and batch sampling.

\subsection{Original pipeline (reproduced)}
\label{sec:orig-pipeline}

\textbf{Stage 1: Activation extraction.} Each prompt $x_i$ is passed through
a frozen backbone LLM $\mathcal{M}$, and the final-layer hidden state at the
last token after prompt prefill position is stored: $h_i = \mathcal{M}(x_i)[-1] \in
\mathbb{R}^{d}$, where $d$ is the backbone's hidden dimension (4096 for
\textsc{LLaMA-3.1-8B}). Extraction is implemented with \texttt{nnsight}
\citep{fiottokaufman2024nnsight}, using an extraction batch size of 5 (we also tested a batch size of 100 and found identical activations), a
maximum sequence length of 512 tokens, and extraction seed 42 for the
base run.

\textbf{Stage 2: Probe training.} A 6-layer MLP $f_\theta: \mathbb{R}^d
\rightarrow \{0,1\}$ (Table~\ref{tab:mlp-arch}) is trained with
cross-entropy loss via AdamW \citep{loshchilov2019decoupled}, learning rate
$2.5\times10^{-4}$, weight decay $1\times10^{-2}$, dropout $0.1$, and a
\texttt{ReduceLROnPlateau} learning-rate schedule (factor 0.5, patience 2),
batch size 1024, for 50 epochs, to reproduce the
original hyperparameters exactly.

\begin{table}[t]
  \centering
  \small
  \begin{tabular}{cccc}
    \toprule
    Layer & In & Out & Activation / Reg. \\
    \midrule
    1 & $d$    & 2048 & GELU + Dropout \\
    2 & 2048   & 2048 & GELU + Dropout \\
    3 & 2048   & 512  & GELU + Dropout \\
    4 & 512    & 512  & GELU + Dropout \\
    5 & 512    & 64   & GELU + Dropout \\
    6 & 64     & 2    & Softmax \\
    \bottomrule
  \end{tabular}
  \caption{MLP probe architecture, reproduced from \citet{khatri2026safety}.
  $d$ is the backbone hidden size and varies by model (Table~\ref{tab:models}).}
  \label{tab:mlp-arch}
\end{table}

\subsection{Extension 1: Cross-architecture generalization}
\label{sec:ext-arch}

We apply the pipeline of Section~\ref{sec:orig-pipeline} without
modification to a set of additional open-weight backbone LLMs spanning
different model families, pretraining corpora, alignment procedures, and parameter counts
(Table~\ref{tab:models}). For each backbone we extract final-layer,
last-token activations on the same three datasets and splits as
\citet{khatri2026safety}, and train an MLP probe with the architecture in
Table~\ref{tab:mlp-arch}, adjusting only the input dimension $d$ to match
the backbone's hidden size. All other hyperparameters (learning rate, batch
size, epochs, optimizer) are held fixed to isolate the effect of the
backbone.

\begin{table}[t]
  \centering
  \small
  \begin{tabular}{lccc}
    \toprule
    Model & Params & $d$ & Family \\
    \midrule
    LLaMA-3.1-8B (orig.) & 8B  & 4096 & Llama \\
    Gemma-4-E4B     & 4B  & 2560 & Gemma \\
    Mistral-7B-v0.3 & 7B  & 4096 & Mistral \\
    Qwen2-7B        & 7B  & 3584 & Qwen \\
    \bottomrule
  \end{tabular}
  \caption{Backbones used in the cross-architecture extension. First row is
  the original paper's setting, included for calibration.}
  \label{tab:models}
\end{table}


\subsection{Extension 2: Sensitivity to non-determinism}
\label{sec:ext-seed}

For select (model, dataset) pair we repeat extraction $K{=}5$ times (seeds 42, 75, 456, 789, 1024), varying:
\begin{itemize}
  \item the order in which prompts are batched during extraction (relevant
  when the inference backend uses non-deterministic batched attention
  kernels)
\end{itemize}
Dataset splits are held fixed across seeds so that variance reflects
extraction/training stochasticity rather than data partitioning. Probes are trained on the train set extracted with seed=42. These probes are then evaluated on test seeds with all 5 seed values. We report standard deviation (computed via non-parametric bootstrap over the $K$ runs) for F1 scores.

As a complementary diagnostic, we additionally treat the extracted
activation tensors themselves as the unit of comparison, independent of any
downstream metric. For a few select (model, dataset, seed) triples we compute a
SHA-256 digest over the serialized activation tensor file produced by Stage
1, and compare digests across seeds within the same (model, dataset) pair.
This lets us distinguish two qualitatively different sources of
``no variance'': (i) Stage 1 produces different activations across seeds,
but Stage 2 happens to converge to indistinguishable probes regardless; or
(ii) Stage 1 itself is seed-invariant, so Stage 2 never sees different
inputs in the first place. Section~\ref{sec:disc-determinism} reports the
result of this check and discusses its implications.

\section{Experimental Setup}
\label{sec:setup}

\paragraph{Datasets.}
We evaluate on three datasets: WildJailbreak (vanilla subset),
BeaverTails (QA pairs with binary \texttt{is\_safe} labels), and
AEGIS~2.0.
For WildJailbreak and AEGIS~2.0 we use an 80/10/10 train/validation/test
split; for BeaverTails we adopt the standard split provided with the dataset.
Table~\ref{tab:datasets} summarises the datasets.

\begin{table}[t]
  \centering
  \tiny
  \setlength{\tabcolsep}{3.5pt}
  \begin{tabular}{lrrrrll}
    \toprule
    Dataset & Train & Val & Test & Total & Label Col & Values \\
    \midrule
    WildJailbreak & 209,247 & 26,156 & 26,156 & 261,559 & \texttt{data\_type} & 0/1 \\
    BeaverTails   & 267,170 & 33,396 & 33,397 & 333,963 & \texttt{is\_safe} & F/T \\
    AEGIS 2.0     & 30,007 & 1,445 & 1,964 & 33,416 & \texttt{label} & safe/unsafe \\
    \bottomrule
  \end{tabular}
  \caption{Datasets used in our experiments with train/validation/test splits, total size, label column, and label values.}
  \label{tab:datasets}
\end{table}

\paragraph{Compute and reproducibility artifacts.} 
We release code, extracted activations and  trained probe checkpoints at \url{https://anonymous.4open.science/\allowbreak r/\allowbreak latent-space-probes-\allowbreak reproducibility-paper-CAFC/}

\paragraph{Evaluation.} We report F1 scores on the held-out
test split of each dataset, matching the original paper's protocol.

\section{Results}
\label{sec:results}

\subsection{Reproduction of the original result}

Table~\ref{tab:repro} compares our reproduced \textsc{LLaMA-3.1-8B}
numbers against those reported by \citet{khatri2026safety}. We find that we are able to reproduce results within a delta of 0.37\% F1 of the values reported in the original work.  

\begin{table}[t]
  \centering
  \small
  \begin{tabular}{lccc}
    \toprule
    Dataset & Orig.\ F1 & Repro.\ F1 & $\Delta$ \\
    \midrule
    WildJailbreak & 99.1 & 99.47 & 0.37 \\
    BeaverTails   & 82.7 & 82.90 & 0.2 \\
    AEGIS 2.0     & 83.5 & 83.42 & -0.08 \\
    \bottomrule
  \end{tabular}
  \caption{Reproduction of \citet{khatri2026safety}, Table 2, using
  identical hyperparameters and splits.}
  \label{tab:repro}
\end{table}

\subsection{Cross-architecture results}

Table~\ref{tab:crossarch} reports probe F1 score for each backbone in
Table~\ref{tab:models} on each dataset.
We find that the probes' performance on the newly introduced model architectures is comparable to their performance to the values reported in the original work. We observe an increase of about 0.4 percentage points in F1 on the BeaverTails dataset for the new models compared to the numbers on LLaMA-3.1-8B.

\begin{table}[t]
  \centering
  \small
  \begin{tabular}{lccc}
    \toprule
    Model & WildJailbreak & BeaverTails & AEGIS 2.0 \\
    \midrule
    LLaMA-3.1-8B  & 99.1 & 82.9 & 83.5 \\
    Gemma-4-E4B   & 98.83 & 83.35 & 83.61 \\
    Mistral-7B-v0.3 & 99.2 & 83.32 & 83.39 \\
    Qwen2-7B       & 99.31 & 83.39 & 83.42 \\
    \bottomrule
  \end{tabular}
  \caption{F1 score across backbones and datasets. First row reproduces the
  original paper's setting.}
  \label{tab:crossarch}
\end{table}

\subsection{Effect of non-determinism}
\label{sec:results-seed}

Table~\ref{tab:perseed} reports the full per-seed breakdown for every
(model, dataset) pair we evaluated under $K{=}5$ extraction seeds. Across every pair, F1 is identical to at least two decimal places across all five seeds: standard deviation is $0$ throughout, and every row of Table~\ref{tab:perseed} is constant across columns. Precision and
recall (not shown in the main table; full values are in Appendix~\ref{sec:appendix-raw}) show the same pattern.

\begin{table}[t]
  \centering
  \tiny
  \setlength{\tabcolsep}{6pt}
  \begin{tabular}{llccccc}
    \toprule
    Model & Dataset & S42 & S75 & S456 & S789 & S1024 \\
    \midrule
    \multirow{2}{*}{LLaMA-3.1-8B} & BeaverTails & 82.90 & 82.90 & 82.90 & 82.90 & 82.90 \\
                                 & AEGIS 2.0   & 83.42 & 83.42 & 83.42 & 83.42 & 83.42 \\
    \addlinespace[1.5ex] 
    Gemma-4-E4B    & BeaverTails & 83.35 & 83.35 & 83.35 & 83.35 & 83.35 \\
    \addlinespace[1.5ex]
    Mistral-7B-v0.3& BeaverTails & 83.32 & 83.32 & 83.32 & 83.32 & 83.32 \\
    \addlinespace[1.5ex]
    Qwen2-7B       & BeaverTails & 83.39 & 83.39 & 83.39 & 83.39 & 83.39 \\
    \bottomrule
  \end{tabular}
  \caption{Per-seed test-set F1 score for every (model, dataset) pair we evaluated under \(K{=}5\) extraction seeds. Columns S42--S1024 denote the five seed values. Every row is constant across columns to the precision shown; see Appendix~\ref{sec:appendix-raw} for precision/recall.(Table~\ref{tab:crossarch}).}
  \label{tab:perseed}
\end{table}

\subsection{Latent state determinism: a hash-matching diagnostic}
\label{sec:disc-determinism}

Given the zero-variance F1 pattern in Section~\ref{sec:results-seed},
we tested whether the extracted activation tensors from Stage 1 were
themselves identical across seeds, rather than merely producing
indistinguishable downstream probes. For each (model, dataset) pair we
computed a SHA-256 digest of the serialized activation tensor file
produced under each of the five extraction seeds and compared digests
pairwise against the seed-42 run. Table~\ref{tab:hashes} reports the
result: for all four model architectures, the digests are identical across
all five seeds for every dataset we checked. In other words, the latent
state tensors extracted from LLaMA-3.1-8B, Gemma-4-E4B, Mistral-7B-v0.3,
and Qwen2-7B are byte-for-byte identical regardless of which of the four
non-default seed values (75, 456, 789, 1024) we compared against the
seed-42 baseline.

\begin{table}[t]
  \centering
  \scriptsize
  \setlength{\tabcolsep}{10pt}
  \begin{tabular}{llcc}
    \toprule
    Model & Dataset & SHA-256 & Match \\
    \midrule
    LLaMA-3.1-8B    & BeaverTails & \texttt{802be5b9} & 4/4 \\
    LLaMA-3.1-8B    & AEGIS 2.0   & \texttt{eff8da81} & 4/4 \\
    Gemma-4-E4B     & BeaverTails & \texttt{e5ce495d} & 4/4 \\
    Mistral-7B-v0.3 & BeaverTails & \texttt{5640ae1e} & 4/4 \\
    Qwen2-7B        & BeaverTails & \texttt{75feb821} & 4/4 \\
    \bottomrule
  \end{tabular}
  \caption{SHA-256 digests (first 8 hex chars) of the serialized Stage-1
  activation tensor file for each (model, dataset) pair. ``Match'' counts,
  out of the four non-default seeds (75/456/789/1024), how many produced
  a digest identical to the seed-42 run.}
  \label{tab:hashes}
\end{table}

\section{Discussion}
\label{sec:discussion}

\textbf{Cross-architecture generality.} The results in
Table~\ref{tab:crossarch} are consistent with the hypothesis in
Section~\ref{sec:ext-arch}: probe F1 score on WildJailbreak and BeaverTails
stays within roughly half a point of the original LLaMA-3.1-8B numbers for
all three additional backbones (Gemma-4-E4B, Mistral-7B-v0.3, Qwen2-7B),
despite these models differing in pretraining corpus, alignment recipe,
and hidden dimensionality (see Appendix). This is consistent with the claim in
\citet{khatri2026safety, saglam2025large, zhao2025latentguard} that safety-relevant information is linearly
decoded in final-layer, last-token representations being a property of
how current aligned LLMs represent language in general, rather than an
artifact specific to the Llama alignment pipeline. We caution, however,
that all four backbones we have tested to date are open-weight,
English-centric, RLHF- or DPO-aligned chat models in a similar 4B--8B
parameter range; we have not yet tested whether the pattern holds at very
different scales, for base (non-aligned) models, or for models aligned with
substantially different objectives.

\textbf{Non-determinism and the guard-model comparison.} Different sub-stages of an LLM behave differently with respect to seed sensitivity, and it is important to keep them separate when interpreting the zero-variance result. From our results, we observe that activation vectors extracted at the last hidden layer of the last prefill token remained invariant to seed variations. This is not the same as claiming that the LLMs generated identical tokens for different seed values.

Non-determinism in LLM inference can be introduced due to various factors \cite{yuan2026understanding}. While a majority of the factors causing non-deterministic outcomes come into play after the first token has been generated, it remains possible for non-deterministic outcomes to occur at the prefill stage, e.g use of non-deterministic algorithms like Chunked Prefill. \cite{agrawal2025efficient}. We interpret the data to infer that it is likely that our current experimental settings were inadequate at recreating non-deterministic behavior in the LLMs across several inference runs. As a result, the complete relationship between non-deterministic LLM behavior and probe performance remains an open question. However, we show that prefill stage activations remain unchanged to batch order despite changes in seed values. 

As Table~\ref{tab:hashes}
confirms, activation extraction, under current experimental settings, is a pure function of the frozen weights
and input tokens. If the variance in F1 patterns were to survive non-determinism in latent activation, it
would suggest that the probe F1 scores are robust to noise in latent space activations, strengthening the original contribution. We do not yet draw
this conclusion. The defensible claim is narrower: our protocol has not
detected variance attributable to minibatch order at the last token last layer level.
This is weaker than a general claim that the probe is variance-free across different implementations.

\textbf{Practical implications.} Taken together, the cross-architecture
and non-determinism results are cautiously encouraging for latent-space
probing as a lightweight alternative or supplement to external guardrails: performance
looks portable across at least a handful of open-weight models, across modalities, and
post-prefill last token activations appear deterministic with respect to variations in seed and batch order. 

\section{Limitations}
\label{sec:limitations}

As with the original paper, we extract only final-layer, last-token
activations; earlier layers or alternative pooling strategies may behave
differently and are left to future work. Concurrent work diagnoses
exactly this failure mode directly: \citet{lasttoken2026} show that
final-token safety probes can miss jailbreak prompts whose unsafe evidence
is distributed across earlier user-token representations rather than
concentrated at the final prefill position, and that naive fixes (wider
probe bottlenecks, naive max-pooling over tokens) do not reliably resolve
the gap.
Our reproduction inherits this limitation
by design, since faithfully reproducing \citet{khatri2026safety} requires
keeping the extraction configuration unchanged. Our non-determinism
analysis (Section~\ref{sec:disc-determinism}) found that extractions of the last hidden state of the last token was deterministic under our experimental settings, confirmed by the digest matches in Table~\ref{tab:hashes}. The zero-variance F1 result should be read as ``no variance detected by our current protocol''
rather than as a general claim about probe stability, and should not be used to argue that single-seed point estimates are reliable outside our specific setup. Our
cross-architecture extension is limited to the backbones in Table~\ref{tab:models} and does not cover closed-weight models, non-English data, or multimodal inputs. Our non-determinism analysis captures extraction- and training-time stochasticity under fixed data splits; it does not capture variance from alternative train/validation/test partitions, which would require a separate cross-validation study. Finally, both the original paper and this reproduction restrict evaluation to prompt-level binary harmfulness classification; findings may not transfer to response-level moderation or
finer-grained, multi-category risk scoring.

\section*{Ethics Statement}

This work analyzes existing safety classification datasets and does not
collect new human data. As with any safety-classifier research, probes of
the kind studied here could in principle be repurposed for overly broad or
opaque content filtering if deployed without transparency; we do not
release any capability beyond what is already available in the original
paper's public artifacts and the datasets' original releases. We do not
reproduce or redistribute any redacted or sensitive content from AEGIS~2.0
beyond what the dataset's original terms permit.

\bibliography{references}
\appendix
\section{Additional Implementation Details}
\label{sec:appendix-impl}
\emph{Extraction Hardware used}: Nvidia T4 (Azure VMs), Nvidia L4 (Google Colab)

Table~\ref{tab:hyperparams} lists the training and extraction
hyperparameters used throughout, matching Section~\ref{sec:orig-pipeline}
and held fixed across all backbones. 

Table~\ref{tab:probeparams} reports
the exact parameter count of the MLP probe for each backbone, following
directly from the fixed architecture in Table~\ref{tab:mlp-arch} and each
backbone's hidden dimension $d$ (Table~\ref{tab:models}).

\begin{table}[h]
  \centering
  \small
  \begin{tabular}{ll}
    \toprule
    \multicolumn{2}{l}{\textit{Training}} \\
    \midrule
    Optimizer & AdamW \\
    Learning rate & $2.5\times10^{-4}$ \\
    Weight decay & $1\times10^{-2}$ \\
    Dropout & 0.1 \\
    Scheduler & ReduceLROnPlateau \\
    Batch size & 1024 \\
    Epochs & 50 \\
    \midrule
    \multicolumn{2}{l}{\textit{Extraction}} \\
    \midrule
    Batch size & 5 \\
    Max sequence length & 512 \\
    Base seed & 42 \\
    \bottomrule
  \end{tabular}
  \caption{Training and extraction hyperparameters, held fixed across all
  backbones and datasets.}
  \label{tab:hyperparams}
\end{table}

\begin{table}[h]
  \centering
  \small
  \begin{tabular}{lcc}
    \toprule
    Model & $d$ & Probe Params \\
    \midrule
    LLaMA-3.1-8B    & 4096 & 13.93M \\
    Gemma-4-E4B     & 2560 & 10.79M \\
    Mistral-7B-v0.3 & 4096 & 13.93M \\
    Qwen2-7B        & 3584 & 12.88M \\
    \bottomrule
  \end{tabular}
  \caption{Exact parameter count of the MLP probe for each backbone,
  determined by the fixed architecture in Table~\ref{tab:mlp-arch} and the
  backbone's hidden dimension.}
  \label{tab:probeparams}
\end{table}
\paragraph{Architectural diversity.}
Table~\ref{tab:arch} summarizes the architectural diversity across our four
backbone models. Despite significant differences in hidden dimensionality
(2560--4096), depth (28--42 layers), attention mechanism (full vs.\
interleaved sliding-window in Gemma-4-E4B), KV head counts, and modality
(Gemma-4-E4B is natively multimodal with text, image, and audio inputs,
while the remaining three are text-only), our MLP probes achieve consistent
F1 score across all backbones (Table~\ref{tab:crossarch}). This suggests the
safety-relevant signal captured by latent-space probes is not tied to any
specific architectural choice.

\begin{table*}[h]
  \centering
  \footnotesize
  \setlength{\tabcolsep}{4pt}
  \begin{tabular}{lcccccl}
    \toprule
    Model & $d$ & Layers & Heads (Q/KV) & Attention & Modality & Arch \\
    \midrule
    LLaMA-3.1-8B    & 4096 & 32 & 32/8 & GQA, full         & Text             & Dense decoder \\
    Gemma-4-E4B     & 2560 & 42 & 8/2  & GQA, sliding+full & Text+Image+Audio & Dense (PLE)   \\
    Mistral-7B-v0.3 & 4096 & 32 & 32/8 & GQA, full         & Text             & Dense decoder \\
    Qwen2-7B        & 3584 & 28 & 28/4 & GQA, full         & Text             & Dense decoder \\
    \bottomrule
  \end{tabular}
  \caption{Architectural summary of the four backbone models. All four use
  RMSNorm. Gemma-4-E4B uses interleaved sliding-window (window~512) and
  full-attention layers (1 full per 6 sliding), per-layer embeddings (PLE),
  and is natively multimodal; the other three are text-only dense decoders.}
  \label{tab:arch}
\end{table*}

\section{Per-Seed Raw Results}
\label{sec:appendix-raw}
The experiment results reported in Table~\ref{tab:seed_metrics} focus on testing variance in F1 with respect to seed. They might not use the most efficient probe model checkpoint, and thus F1 values may differ slightly from those reported in the main paper.

\begin{table*}[htbp]
\centering
\footnotesize
\setlength{\tabcolsep}{6pt}
\begin{tabular}{llrrrr}
\toprule
Model & Dataset & Seed & Precision & Recall & F1 \\
\midrule
Gemma   & BeaverTails & 1024 & 0.8346 & 0.8249 & 0.8297 \\
        &             & 456  & 0.8346 & 0.8249 & 0.8297 \\
        &             & 75   & 0.8346 & 0.8249 & 0.8297 \\
        &             & 789  & 0.8346 & 0.8249 & 0.8297 \\
\midrule
Llama   & AEGIS       & 1024 & 0.8229 & 0.8470 & 0.8348 \\
        &             & 456  & 0.8229 & 0.8470 & 0.8348 \\
        &             & 75   & 0.8229 & 0.8470 & 0.8348 \\
        &             & 789  & 0.8229 & 0.8470 & 0.8348 \\
\midrule
Llama   & BeaverTails & 1024 & 0.8554 & 0.7965 & 0.8249 \\
        &             & 456  & 0.8554 & 0.7965 & 0.8249 \\
        &             & 75   & 0.8554 & 0.7965 & 0.8249 \\
        &             & 789  & 0.8554 & 0.7965 & 0.8249 \\
\midrule
Mistral & BeaverTails & 1024 & 0.8360 & 0.8226 & 0.8292 \\
        &             & 456  & 0.8360 & 0.8226 & 0.8292 \\
        &             & 75   & 0.8360 & 0.8226 & 0.8292 \\
        &             & 789  & 0.8360 & 0.8226 & 0.8292 \\
\midrule
Qwen    & BeaverTails & 1024 & 0.8355 & 0.8247 & 0.8301 \\
        &             & 456  & 0.8355 & 0.8247 & 0.8301 \\
        &             & 789  & 0.8355 & 0.8247 & 0.8301 \\
\bottomrule
\end{tabular}
\caption{Precision, recall, and F1 for each model / dataset / seed triplet}
\label{tab:seed_metrics}
\end{table*}
\end{document}